%% file: main.tex
\pdfoutput=1

\documentclass[11pt]{article}

\usepackage{ACL2023}
\usepackage{times}
\usepackage{latexsym}

\usepackage[T1]{fontenc}
\usepackage[utf8]{inputenc}
\usepackage{microtype}
\usepackage{inconsolata}
\usepackage{algorithm}
\usepackage{algorithmic}
\usepackage{makecell}
\usepackage{multirow}
\usepackage{CJKutf8}
\usepackage{color}
\usepackage{pgfplots}
\usepackage[normalem]{ulem}
\usepackage{graphicx}
\usepackage{amsfonts}
\usepackage{amsmath,amssymb,bm}
\usepackage{times}
\usepackage{latexsym}
\usepackage{color,xcolor}
\usepackage{CJKutf8}
\usepackage{amsmath,amsfonts}
\usepackage{algorithmic}
\usepackage{array}
\usepackage[caption=false,font=normalsize,
   labelfont=sf,textfont=sf]{subfig}
\usepackage{textcomp}
\usepackage{stfloats}
\usepackage{url}
\usepackage{verbatim}
\usepackage{graphicx}
\usepackage{balance}
\usepackage{inconsolata}
\usepackage{booktabs}
\usepackage{paralist}
\usepackage{colortbl}
\usepackage{algorithm}
\usepackage{algorithmic}
\usepackage{makecell}
\usepackage{arydshln}
\usepackage{multirow}
\usepackage{color}
\usepackage{pgfplots}
\usepackage[normalem]{ulem}
\usepackage{graphicx}
\usepackage{amsfonts}
\usepackage{amsmath,amssymb,bm}
\usepackage{soul}
\usepackage{amsmath}
\usepackage{tabularx}
\usepackage{booktabs}
\usepackage{amssymb}
\usepackage{pifont}
\usepackage{bm}
\usepackage{multirow}
\usepackage{tcolorbox}
\usepackage[framemethod=tikz]{mdframed}
\usepackage{tikz,pgfplots}
\usepgfplotslibrary{groupplots}
\pgfplotsset{compat=1.13}

\definecolor{nmgray}{RGB}{229,229,229}
\definecolor{underlinegray}{RGB}{197,197,197}

\definecolor{introblue}{RGB}{0,176,240}
\definecolor{introgreen}{RGB}{0,203,134}
\definecolor{introgreen2}{RGB}{139,243,206}

\usepackage{tcolorbox}

\usepackage{adjustbox}

\newtcolorbox{mybox}[2][]{
width=\columnwidth,
colback = nmgray!75!white, 
colframe = nmgray!75!white, 
boxsep=0pt,left=10pt,right=10pt,top=0pt,bottom=0pt,
fontupper=\linespread{0.9}\selectfont,
title=#2,#1}

\newcommand{\circleone}[1]{%
    \resizebox{!}{0.8em}{%
        \tikz[baseline=(char.base)]{
            \node[shape=circle, fill=black, inner sep=0.8pt, text=white] (char) {#1};
        }%
    }%
}

\newcommand{\circletwo}[1]{%
    \resizebox{!}{0.8em}{%
        \tikz[baseline=(char.base)]{
            \node[shape=circle, fill=black, inner sep=0.8pt, text=white] (char) {#1};
        }%
    }%
}

\newcommand{\circlethree}[1]{%
    \resizebox{!}{0.8em}{%
        \tikz[baseline=(char.base)]{
            \node[shape=circle, fill=black, inner sep=0.8pt, text=white] (char) {#1};
        }%
    }%
}

\usepackage{CJKutf8}
\usepackage{color}

\title{Enhancing Hyperbole and Metaphor Detection with Their \\ Bidirectional Dynamic Interaction and Emotion Knowledge}

\author{
  Li Zheng\textsuperscript{\rm 1},
 Sihang Wang\textsuperscript{\rm 1},
    Hao Fei\textsuperscript{\rm 2},
    Zuquan Peng\textsuperscript{\rm 1},
    Fei Li\textsuperscript{\rm 1}\thanks{     
    $\,$ Corresponding author.},
    Jianming Fu\textsuperscript{\rm 1}, \\
    \textbf{Chong Teng\textsuperscript{\rm 1}},
    \textbf{Donghong Ji\textsuperscript{\rm 1}}
  \\
  \textsuperscript{\rm 1}Key Laboratory of Aerospace Information Security and Trusted Computing, Ministry of \\ Education, School of Cyber Science and Engineering, Wuhan University, Wuhan, China\\
  \textsuperscript{\rm 2}National University of Singapore, Singapore, Singapore
  \\
\texttt{\{zhengli,sihangwang,pzq\_cse,lifei\_csnlp,jmfu,tengchong,dhji\}@whu.edu.cn} \\ 
    \texttt{haofei37@nus.edu.sg}}

\begin{document}
\begin{CJK}{UTF8}{gbsn}
\maketitle

\input{Section/0-abstract}
\input{Section/1-introduction}

\input{Section/2-relatedwork}

\input{Section/3-methodology}

\input{Section/4-experiments}
\input{Section/5-conclusion}

\bibliography{main}
\bibliographystyle{acl_natbib}

\end{CJK}
\end{document}

%% file: Section/0-abstract.tex
\begin{abstract}

Text-based hyperbole and metaphor detection are of great significance for natural language processing (NLP) tasks. 
However, due to their semantic obscurity and expressive diversity, it is rather challenging to identify them. 
Existing methods mostly focus on superficial text features, ignoring the associations of hyperbole and metaphor as well as the effect of implicit emotion on perceiving these rhetorical devices.
To implement these hypotheses, 
we propose an emotion-guided hyperbole and metaphor detection framework based on bidirectional dynamic interaction (EmoBi).
Firstly, the emotion analysis module deeply mines the emotion connotations behind hyperbole and metaphor. 
Next, the emotion-based domain mapping module identifies the target and source domains to gain a deeper understanding of the implicit meanings of hyperbole and metaphor. 
Finally, the bidirectional dynamic interaction module enables the mutual promotion between hyperbole and metaphor. 
Meanwhile, a verification mechanism is designed to ensure detection accuracy and reliability.
Experiments show that EmoBi outperforms all baseline methods on four datasets.
Specifically, compared to the current SoTA, the F1 score increased by 28.1\% for hyperbole detection on the TroFi dataset and 23.1\% for metaphor detection on the HYPO-L dataset.
These results, underpinned by in-depth analyses, underscore the effectiveness and potential of our approach for advancing hyperbole and metaphor detection.

\end{abstract}

%% file: Section/1-introduction.tex
\section{Introduction}

Hyperbole and metaphor, as common rhetorical devices, not only enrich language expressions but also play a crucial role in emotion conveyance \cite{mohammad2016metaphor,djokic2021metavr} and semantic understanding \cite{neuman2013metaphor,ding2025zero-shot,he2025dalr}.
Therefore, the accurate detection and understanding of hyperbole and metaphor are of critical significance for improving the performance of many Natural Language Processing (NLP) tasks, such as emotion analysis systems \cite{zheng2023bi,zhang2024camel,lee2023mindterior} and intelligent chatbots \cite{samad2022empathetic,zheng2023ecqed,xie2024does,zheng2025multi}. 
However, due to their semantic obscurity and expressive diversity, identifying hyperbole and metaphor has always been a challenging issue in NLP research.

\begin{figure}[!t]
    \centering
    \includegraphics[scale=0.44]{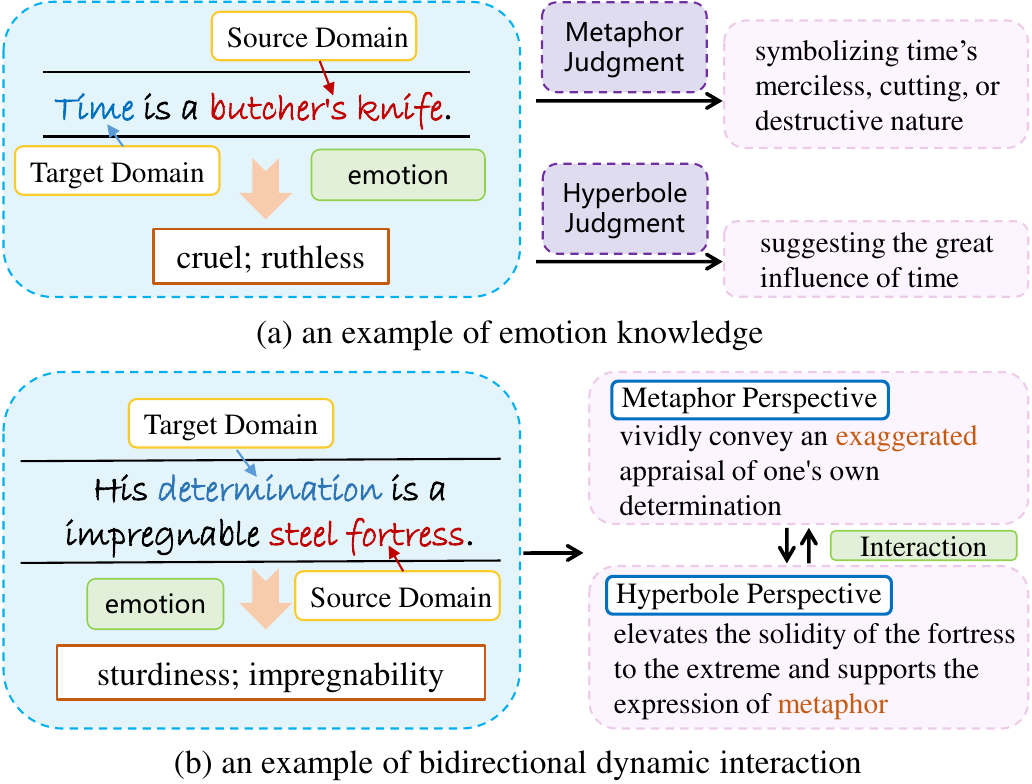}
    \caption{Examples of Hyperbole and Metaphor Detection.
    }
    \label{fig:ex}
    \vspace{-0.6cm}
\end{figure}

Several studies have made commendable efforts in hyperbole and metaphor detection. 
Some researches build separate detection models for  hyperboles \cite{tian2021hypogen,schneidermann2023probing} or metaphors \cite{tian2024bridging,zhang2024gome}. 
Additionally, the latest work \cite{badathala2023match} considering the interaction between hyperboles and metaphors, proposed a multi-task method for simultaneously detecting them.
However, these methods mainly focus on the extraction of surface-level text features and implicit feature sharing among tasks. They ignore the emotions behind rhetorical expressions and the dynamic interaction between tasks.
Specifically, the employment of rhetorical devices is often emotion-driven \cite{dankers2019modelling,chen2023identifying,zhang2024camel}, and different devices can interact and jointly construct semantics. 
Therefore, \textbf{(1) how to mine the emotions behind hyperbole and metaphor} and \textbf{(2) how to model the dynamic interaction relationship between them}, and utilize these to achieve detection are of crucial importance.

On the one hand, most prior approaches \cite{elzohbi2023contrastwsd,zhang2023image,badathala2023match} only focus on lexical and syntactic features, while the consideration of emotion factors remains insufficient. 
In fact, emotion is the vehicle of semantic expression and a key factor in facilitating the understanding of rhetorical effects.
As shown in Figure \ref{fig:ex} (a), from the emotion perspective, the term ``butcher's knife'' carries a cruel and ruthless emotion connotation. 
Without emotion knowledge, it is arduous to fathom that ``time'' (the target domain) is metaphorically referred to as a ``butcher's knife'' (the source domain) just from a literal interpretation. 
It might be misinterpreted as a description of an actual knife.
For hyperbole detection, the cruel and ruthless ``butcher's knife'' holds a hyperbolic significance, intimating the great influence of time.
Absent the aid of emotion knowledge, it could be misapprehended as a meaningless text.

On the other hand, although hyperboles and metaphors differ in linguistic manifestations, they possess certain inherent associations as both involve deviations from the literal meaning to achieve specific expressive effects \cite{carston2011metaphor,burgers2016figurative}. 
Nevertheless, existing methods \cite{troiano2018computational,badathala2023match,qiao2024quantum} either treat these two rhetorical devices separately or simply conduct implicit feature fusion, neglecting to explicitly explore and utilize the bidirectional dynamic interaction process between them.
As illustrated in Figure \ref{fig:ex} (b), from the metaphor perspective, comparing ``determination'' to ``steel fortress'' vividly exaggerates the firmness of determination via the fortress's sturdiness and impregnability. 
Conversely, this hyperbole elevates the sturdiness of the fortress to the extreme, reinforcing the metaphorical link between ``determination'' and ``steel fortress''.

Based on the above observations, we propose an \textit{\underline{\textbf{Emo}}tion-guided hyperbole and metaphor detection framework based on \underline{\textbf{Bi}}directional Dynamic Interaction (EmoBi)}.
\textbf{Firstly}, we conduct an emotion analysis of the sentence. 
By excavating the deep-seated correlations between emotions and hyperboles as well as metaphors within the text, crucial cues are provided for subsequent identification and comprehension.
\textbf{Secondly}, we perform emotion-based domain mapping. 
Based on the emotion analysis results, we prompt the large language model (LLM) to identify the target domain and the source domain from the emotion perspective.
This enriches the semantic representation of the target domain through emotional connotations, facilitating a deeper understanding of implicit meaning in hyperboles and metaphors.
\textbf{Finally}, we design a bidirectional dynamic interaction mechanism to enable hyperbole and metaphor to mutually reinforce each other.
The intense emotion and degree variation within hyperbole render the conceptual mapping of metaphor more profound and expressive. 
Meanwhile, metaphor sets the semantic framework and emotional tone for hyperbole.
Additionally, we set up a verification mechanism to ensure detection accuracy and reliability.

To validate the effectiveness of our model, we conduct experiments on four widely used datasets for hyperbole and metaphor detection, namely HYPO \cite{troiano2018computational}, and HYPO-L \cite{zhang2021mover}, LCC \cite{mohler2016introducing}, TroFi \cite{birke2006clustering}. 
The experimental results show that our model significantly outperforms all state-of-the-art (SoTA) baselines on all evaluation metrics. 
Specifically, in hyperbole and metaphor detection, the F1 scores are improved by 28.1\% on the TroFi dataset and 23.1\% on the HYPO-L dataset respectively compared to the current SoTA.
Moreover, we carry out a large number of experiments to show the effectiveness of the emotion guidance and the bidirectional dynamic interaction. 
Our main contributions are summarized as follows:
\begin{itemize}
\item 
We propose a novel emotion-guided framework to understand hyperbole and metaphor comprehensively through emotion expressions, providing a new perspective for rhetorical language study.

\item 
We design a bidirectional dynamic interaction mechanism that promotes the mutual enhancement between hyperboles and metaphors.

\item 
Our extensive experimental results on the four widely used hyperbole and metaphor datasets demonstrate that our scheme achieves SoTA performance.\footnote{Our codes: https://github.com/ZhengL00/EmoBi.}
\end{itemize}

%% file: Section/2-relatedwork.tex
\section{Related Work}

\subsection{Hyperbole and Metaphors Detection}

Hyperbole and metaphor detection has been an active research area in Natural Language Processing (NLP) \cite{zhang2024image,kalarani2024unveiling,govindan2022machine}.
For metaphor detection, \citet{birke2006clustering} developed the TroFi dataset, focusing on literal and metaphorical verb usages. 
\citet{mohler2016introducing} contributed the LCC dataset with sentence-level metaphor annotations in four languages. 
\citet{tian2024bridging} proposed a domain mining method based on interpretable word pairs for metaphor detection. 
\citet{yang2024can} bootstraped and combined tacit knowledge to conduct verb metaphor detection.
In the realm of hyperbole detection, \citet{mccarthy2004there} established a theoretical framework, which provided a foundation for subsequent research efforts.
\citet{troiano2018computational} took a significant step forward by developing the first comprehensive hyperbole dataset.
\citet{tian2021hypogen} utilized common sense and counterfactual knowledge to generate sentence-level hyperboles. 
\citet{schneidermann2023probing} explored hyperbole detection in pre-trained language models.
Nevertheless, these methods typically handle metaphors or hyperboles independently, overlooking the interactions between them. 
\citet{badathala2023match} proposed a multi-task approach that considered the mutual promotion between hyperboles and metaphors. 
However, they mainly focused on surface feature sharing and insufficiently considered the emotion guidance and the deep-level interaction between hyperboles and metaphors.

\subsection{Emotion Analysis}

Significant progress has been made in emotion analysis \cite{liu2020sentiment,akhtar2016hybrid,pang2002thumbs,zheng2024reverse,zheng2024self}, evolving from early dictionary matching methods to deep learning models.
\citet{turney2002thumbs} proposed a pointwise mutual information measure method for predicting emotions.
With the emergence of deep learning, \citet {rakhlin2016convolutional} introduced convolutional neural networks for sentiment analysis.
\citet{wang2023emp} found that emotions can be effectively used to study personality traits.
Regarding the research on the association between emotions and rhetoric.
\citet{mohammad2016metaphor} explored the relationship between metaphor and emotion, finding that metaphors often carry emotion tendencies. 
\citet{chen2023identifying} proposed an emotion recognition model through hierarchical structure and rhetorical correlation.
Considering the driving role of emotion on rhetorical devices, we propose an emotion-guided framework to achieve more accurate and comprehensive detection.

\subsection{LLM Reasoning}

The emergence of large language models (LLMs) has opened up new avenues for hyperbole and metaphor detection \cite{mann2020language,wu2023next,xu2024exploring}. 
Prompting techniques have been widely explored to elicit the knowledge and reasoning capabilities of LLMs \cite{li2023unified,liang2024multi,zhou2024token}.
Chain-of-Thought (CoT) prompting and its variants have been proposed to guide LLMs in generating intermediate reasoning steps, which have shown promising results in improving performance on complex tasks \cite{wei2022chain,yao2024tree,besta2024graph}.
Although existing prompting techniques have achieved success in other domains, the unique characteristics of hyperbole and metaphor, such as their implicitness and context-dependence, pose additional challenges. 
There is a lack of systematic research to effectively prompt LLMs to capture the essence of these rhetorical devices. 
Our work builds upon these previous efforts and proposes a novel emotion-guided framework that explicitly models the emotion knowledge and dynamic interaction between hyperboles and metaphors.

%% file: Section/3-methodology.tex
\section{Methodology}

\begin{figure*}[!h]
    \centering
    \includegraphics[scale=0.6]{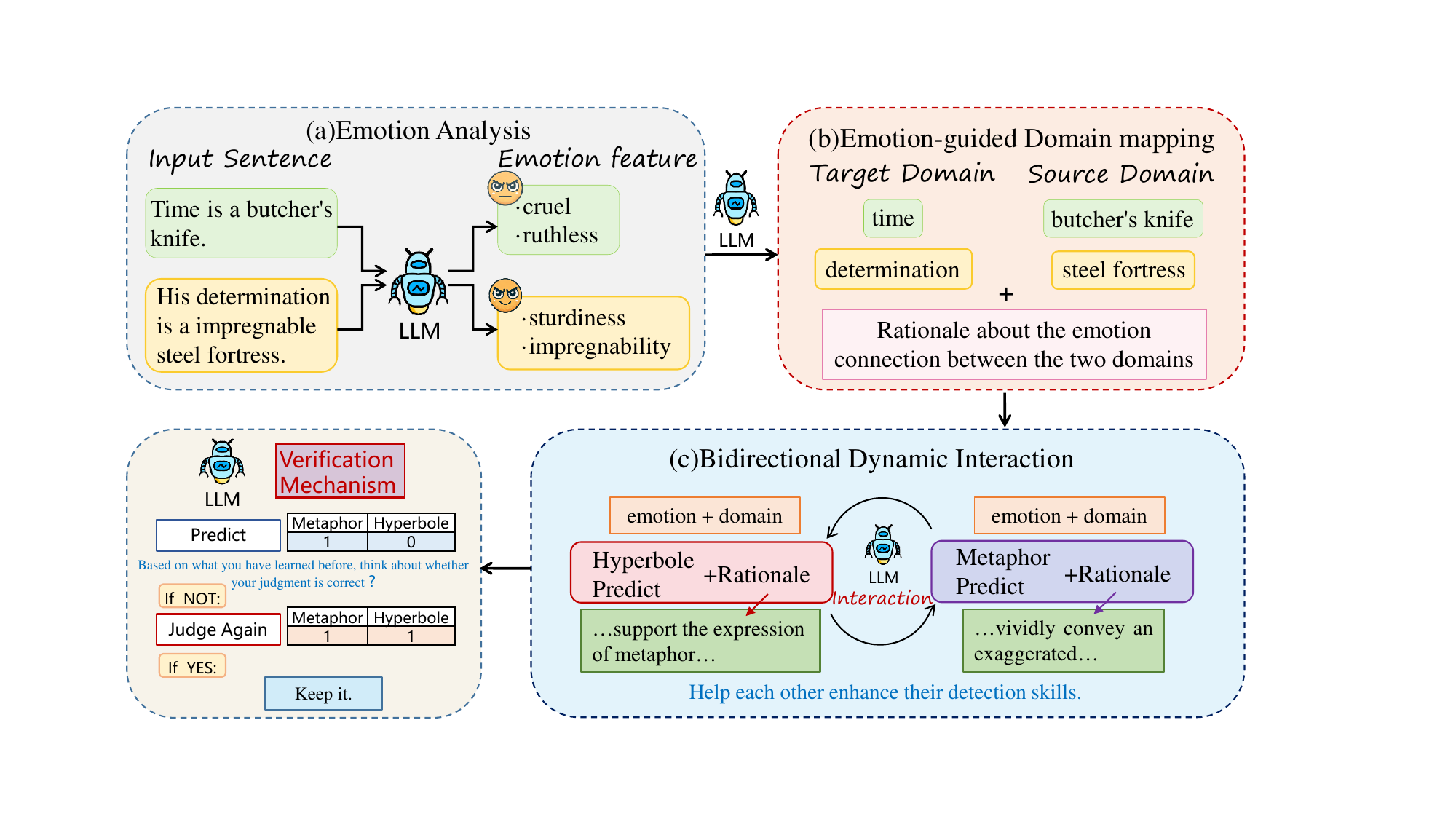}
    \caption{
    The overall architecture of our model.}
    \label{fig:model}
\end{figure*}

\subsection{Task Definition}

For a given sentence x , our goal is to predict the corresponding hyperbole label $y_h$ and metaphor label $y_m$, that is  $f(x)\to (y_h, y_m)$ , where $f(x)$ represents the detection model.

\subsection{Method Overview}

In this paper, we propose an emotion-guided hyperbole and metaphor detection framework based on bidirectional dynamic interaction (EmoBi), which fully utilizes emotion information and inter-task interactions to enhance hyperbole and metaphor detection.
The architecture of our framework is illustrated in Figure \ref{fig:model} and comprises three components: (1) an emotion analysis module, (2) an emotion-guided domain mapping module, and (3) a bidirectional dynamic interaction module. 
The emotion analysis module, captures the emotion context of the sentence, supplying key cues for later detection.
The domain mapping module utilizes semantic and emotion information to identify target and source domains, aiding in understanding implicit meanings. 
Finally, the bidirectional dynamic interaction mechanism enables mutual promotion in detection via knowledge transfer.

\subsection{Emotion Analysis}

The first stage involves a comprehensive sentiment analysis of the given sentence. 
This step is of utmost importance as emotion is a key factor in facilitating the understanding of rhetorical effects. 
By deeply mining the emotion within the sentence, it can effectively connect the surface level of language with the deeper level of rhetoric, thus contributing to a precise interpretation of the rhetorical effects of the text.
Specifically, we prompt the LLM to analyze the emotion of the sentence. 
The specific input template is as follows:
\begin{mybox}\texttt
\texttt{\textbf{Input}: <sentence>}
\end{mybox}
\begin{mybox}\texttt
\texttt{\textbf{Prompt1}: Please analyze the emotion of the following sentence.}
\end{mybox}

This step can be formulated as:
\begin{equation}
x_e = LLM(x, Prompt1)
\end{equation}
where $x_e$ represents the result of emotion analysis.
The LLM processes the input and returns the emotion information. 
Emotion analysis not only enables us to understand the emotion background of hyperboles and metaphors but also provides crucial guidance for subsequent domain mapping.

\subsection{Emotion-Based Domain Mapping}

Domain mapping is of crucial significance as it facilitates the comprehension of semantic transfer and conceptual relationships within a sentence, which are typically the key elements in the  detection of hyperboles and metaphor.
Based on the emotion analysis result from the previous step, we prompt the LLM to perform domain mapping from an emotion perspective and identify the source domain and target domain of the sentence. 
We design the following prompt template for the LLM:
\begin{mybox}\texttt
\texttt{\textbf{Input}: <sentence> + <emotion analysis>}
\end{mybox}
\begin{mybox}\texttt
\texttt{\textbf{Prompt2}: Based on the above emotion analysis result, identify the source domain and target domain in the sentence, and analyze the emotion connection between the two domains.}
\end{mybox}

This step can be formulated as:
\begin{equation}
x_d = LLM(x, x_e, Prompt2)
\end{equation}
where $x_d$ contains the source domain, the target domain, and the corresponding explanations.
The identification of source and target domains constructs a crucial bridge for understanding hyperbole and metaphor. 
The source domain, as the initial conceptual foundation of hyperbolic and metaphorical expressions, bears fundamental semantic features and emotional connotations. 
The target domain is the destination where these semantic features and emotional connotations are transferred and mapped.
By identifying the source domain and the target domain, the starting and ending points of semantic transfer can be accurately located.
This allows for a clear examination of semantic magnification or distortion in hyperbole and cross-domain conceptual mapping and fusion in metaphor. 
Consequently, it promotes a more comprehensive understanding of sentence semantics and more precise detection of hyperbole and metaphor.

\subsection{Bidirectional Dynamic Interaction}

Utilizing the obtained emotion knowledge and domain understanding, we design a bidirectional dynamic interaction mechanism to further perform hyperbole and metaphor detection. 
In this mechanism, hyperbole and metaphor mutually reinforce each other. 
The intense emotions and degree changes inherent in hyperbole can provide richer semantic expansion directions for metaphor, enhancing the depth and expressiveness of the metaphorical concept mapping. 
Conversely, metaphor sets the semantic framework and emotion tone for hyperbole, making the degree changes in hyperbole more reasonable and coherent. 
This bidirectional dynamic interaction promotes mutual learning between the hyperbole and metaphor detection tasks, thereby improving the accuracy and efficiency of detection. 

Taking metaphor-guided hyperbole detection as an example, we prompt the LLM to perform metaphor detection based on the knowledge obtained from the previous two steps, obtaining metaphor detection knowledge. 
\begin{equation}
x_m = LLM(x, x_e, x_d)
\end{equation}
where $x_m$ denotes the metaphor information in the sentence.
Then, based on the prior emotion knowledge, domain knowledge, and metaphor information, we conduct hyperbole detection.

\begin{mybox}\texttt
\texttt{\textbf{Input}: <sentence> + <emotion analysis> + <domain mapping> + <metaphor analysis> }
\end{mybox}
\begin{mybox}\texttt
\texttt{\textbf{Prompt3}: Based on the emotion knowledge, domain knowledge, and metaphor knowledge, analyze whether the sentence is a hyperbole sentence. }
\end{mybox}

This step can be formulated as:
\begin{equation}
y_h = LLM(x, x_e, x_d, x_m, Prompt3)
\end{equation}

The LLM analyzes the sentence based on the provided information and outputs the hyperbole label $y_h$. 
Conversely, the process of hyperbole-guided metaphor detection is similar. 
First, we utilize the emotion and domain knowledge from the previous two steps to analyze the hyperbole information $x_h$ in the sentence.
Subsequently, the final metaphor label $y_m$ is derived based on the emotion knowledge $x_e$, domain knowledge $x_d$, and hyperbole information $x_h$.
This bidirectional dynamic interaction not only improves the detection of each individual rhetorical device but also enriches the overall understanding of the semantic and rhetorical complexity of the text.

Furthermore, we design a validation mechanism. 
If an error is detected in the identified hyperboles or metaphors, the model re-evaluate and adjust the results. 
Through the validation mechanism, we ensure the accuracy and reliability of the hyperbole and metaphor detection and improve the overall performance of the framework.

%% file: Section/4-experiments.tex
\section{Experiments}
\subsection{Experimental Setting}

\noindent\textbf{Datasets.}
We evaluate the effectiveness of our framework on four widely-used datasets with both hyperbole and metaphor labels, namely HYPO \cite{troiano2018computational}, and HYPO-L \cite{zhang2021mover}, LCC \cite{mohler2016introducing}, TroFi \cite{birke2006clustering}.

\noindent\textbf{Evaluation Metrics.} In terms of evaluation metrics, we align with \cite{badathala2023match} and use three metrics, namely precision (P), and recall (R), and F1, to assess the performance.

\subsection{Baseline Systems} 

To verify the effectiveness of our model, we compare it with the following state-of-the-art baselines. 
(1) \citet{badathala2023match} propose a multi-task method with a fully shared layers (MTL-F) model based on BERT \cite{devlin2018bert}, ALBERT \cite{lan2019albert}, and RoBERTa \cite{liu2019roberta} respectively.

\noindent\textbf{(2) Standard Prompting.}
Standard prompting methods have been widely utilized in previous works \cite{ma2023hybridprompt,zhu2024promptbench}. 
For this task, we construct the following prompt template as the input for LLMs:

\begin{mybox}\texttt
\texttt{\textbf{Prompt}: Please identify the hyperbole label $y_h$ and metaphor label $y_m$ of the following sentence $x$.}
\end{mybox}
Nevertheless, this method lacks explicit guidance for the LLM's step-by-step reasoning process, diminishing the interpretability of their answers and making it challenging to understand the underlying logic behind the LLM's responses.

\noindent\textbf{(3) Vanilla CoT Prompting.}
To enhance the standard prompting method, chain-of-thought (CoT) prompting has been investigated \cite{wei2022chain}.
It has made progress not only in generating answers but also in inspiring the LLM to provide the rationale basis behind the answers. 
We construct the following prompt template as inputs to LLMs:
\begin{mybox}\texttt
\texttt{\textbf{Prompt}: Let's think step by step to identify the hyperbole label $y_h$ and metaphor label $y_m$ of the following sentence $x$.}
\end{mybox}

However, the CoT merely prompts the model to directly generate the intermediate reasoning process.
It falls short of delving into the emotion background behind hyperbole and metaphor and the profound interaction between them.

\begin{table*}[htp!]
\centering
\resizebox{\textwidth}{!}{
\renewcommand{\arraystretch}{1}
\begin{tabular}{lccccccccccccccc}
\hline
\multirow{2}[0]{*}{\textbf{Method}} & \multicolumn{3}{c}{\textbf{Hyperbole}} & &\multicolumn{3}{c}{\textbf{Metaphor}} & & \multicolumn{3}{c}{\textbf{Hyperbole}} & &\multicolumn{3}{c}{\textbf{Metaphor}} 
\\
\cline{2-4} \cline{6-8} \cline{10-12} \cline{14-16}

 & \textbf{P} & \textbf{R} & \textbf{F1} & & \textbf{P} & \textbf{R} & \textbf{F1} & &   \textbf{P} & \textbf{R} & \textbf{F1} && \textbf{P} & \textbf{R} & \textbf{F1}
 \\
\hline
 & \multicolumn{7}{c}{\textbf{HYPO}} & &  \multicolumn{7}{c}{\textbf{HYPO-L}} \\
 \cline{2-8}\cline{10-16}
 
MTL-F-BERT & 85.3&82.4 &83.6 &&79.9& 68.6& 72.9&& 65.5& 61.9&63.8 &&55.2 &45.4 &50.3 \\
MTL-F-ALBERT & 84.7 & 87.7&86.0&&75.7 &76.1 &75.3 && 63.8&59.3&61.4&&49.8&38.5&43.0\\
MTL-F-RoBERTa & \textbf{87.9}&88.4 &\underline{88.1} &&\textbf{82.6} &75.2 &\underline{78.7} &&\underline{70.6} &66.8&68.7&&59.9&55.4&57.2\\
Prompt-based &71.6 &90.2 &79.8 && 72.8& 70.1&71.4 &&62.4&77.3&69.1&&61.9&73.2&67.1\\
CoT-based & 76.1&\underline{91.8} & 83.2&& 75.4&\underline{79.2} &77.2 &&67.5&\underline{78.7}&\underline{72.8}&&\underline{65.3}&\underline{81.7}&\underline{72.6}\\
\textbf{Ours} &\underline{87.7}&\textbf{94.1} &\textbf{90.8} &&\underline{81.2} &\textbf{88.1} & \textbf{84.5}&& \textbf{74.2}&\textbf{85.1}&\textbf{79.3}&&\textbf{75.8}&\textbf{85.4}&\textbf{80.3}\\
\specialrule{0em}{-2pt}{-1pt}  
 & \scriptsize{(-0.2\%)} & \scriptsize{(+2.3\%)} & \scriptsize{(+2.7\%)} && \scriptsize{(-1.4\%)} & \scriptsize{(+8.9\%)} & \scriptsize{(+5.8\%)} && \scriptsize{(+3.6\%)} & \scriptsize{(+6.4\%)} & \scriptsize{(+6.5\%)} && \scriptsize{(+10.5\%)} & \scriptsize{(+3.7\%)} & \scriptsize{(+7.7\%)} \\

\hline

& \multicolumn{7}{c}{\textbf{LCC}} & &\multicolumn{7}{c}{\textbf{TroFi}} \\
  \cline{2-8}\cline{10-16}

MTL-F-BERT & 63.3&53.1 &57.5 &&75.0 &77.4 &76.0 && 56.5&43.3&48.6&&55.6&52.5&54.0\\
MTL-F-ALBERT & 61.4&42.5 &49.9 &&70.9 &78.5 &74.4 && 48.7&24.1&31.2&&51.6&45.7&47.5\\
MTL-F-RoBERTa &63.0 &69.1 &65.9 &&79.8 &\underline{81.2} &80.5 && 60.5&52.9&56.1&&56.5&58.7&57.3\\
Prompt-based &61.4 &87.9 &72.3 &&82.3 &69.1 &75.2 &&68.1&79.4&73.3&&82.4&56.3&66.9\\
CoT-based &\underline{68.1} &\underline{90.1} &\underline{77.5} &&\underline{89.4} &78.4 &\underline{83.6} &&\underline{71.3}&\underline{87.3}&\underline{78.5}&&\underline{83.5}&\underline{61.2}&\underline{70.7}\\

\textbf{Ours} &\textbf{76.3} &\textbf{95.6}&\textbf{84.9}&&\textbf{95.7}&\textbf{87.3}&\textbf{91.3}&&\textbf{76.6}&\textbf{93.5}&\textbf{84.2}&&\textbf{91.3}&\textbf{65.9}&\textbf{76.6}\\
\specialrule{0em}{-2pt}{-1pt}   
& \scriptsize{(+8.2\%)} & \scriptsize{(+5.5\%)} & \scriptsize{(+7.4\%)} && \scriptsize{(+6.3\%)} & \scriptsize{(+6.1\%)} & \scriptsize{(+7.7\%)} && \scriptsize{(+5.3\%)} & \scriptsize{(+6.2\%)} & \scriptsize{(+5.7\%)} && \scriptsize{(+7.8\%)}& \scriptsize{(+4.7\%)}& \scriptsize{(+5.9\%)}\\
\hline
\end{tabular}}
\caption{Experimental results on hyperbole and metaphor detection. In the brackets are the improvements of our model over the best-performing baseline(s). MTL-F-RoBERTa is the current SoTA.}
\label{tab:main result}
\vspace{-4mm}
\end{table*}

\subsection{Main Results}

The experimental results of the hyperbole and metaphor detection tasks on four datasets are shown in Table \ref{tab:main result}. 
The results highlight that our method outperforms the SoTA baselines on four datasets, revealing several key findings.
Firstly, compared with prompt-based and CoT-based reasoning, our method has unique advantages. 
On the LCC dataset, the F1 score of the metaphor detection has risen by 16.1\% and 7.7\% respectively compared with prompt-based and CoT-based reasoning, and the F1 score of the hyperbole detection has increased by 12.6\% and 7.4\%. 
This firmly validates that in-depth text emotion analysis and hyperbole-metaphor interaction exploration enable a more precise grasp of their nuanced semantics.
In addition, contrasted with the current SoTA (MTL-F-RoBERTa), our method presents more prominent benefits. 
On the TroFi dataset, the F1 scores of the hyperbole detection has surged by 28.1\%.
And on the HYPO-L dataset, the F1 scores of the metaphor detection has increased by 23.1\%.
This indicates that it is insufficient for MTL-F to mine the clues of hyperbole and metaphor only from the surface features. 
It also shows the necessity and effectiveness of understanding the emotional background behind the sentence. 
The hyperbole providing a richer semantic expansion direction for the metaphor and the metaphor setting a semantic framework for the hyperbole.
Furthermore, compared to the current SoTA, the F1 scores of prompt-based and CoT-based reasoning in hyperbole detection on the HYPO dataset decreased by 8.3\% and 4.9\% respectively, and those in metaphor detection declined by 7.3\% and 1.5\% respectively.
This indicates that relying solely on the inference ability of the LLM itself is inadequate and further demonstrates the effectiveness of our method.

\subsection{Ablation Study}

We perform ablation experiments to evaluate the contribution of each component in our model. 
As depicted in Table \ref{tab:ablation}, no variant matches the full model’s performance, highlighting the indispensability of each component.
Specifically, when the emotional analysis is not utilized, the performance degradation is the most prominent across both tasks on all four datasets.
Particularly, on the metaphor detection of the HYPO-L dataset, the F1 score dropped by 5.7\%. 
This indicates the significance of providing emotional context.
To verify the necessity of the bidirectional dynamic interaction mechanism, we remove this module. 
The sharp decline in the results demonstrate that this module enables a mutual reinforcement between hyperbole and metaphor, facilitating a more comprehensive and accurate understanding of their semantic relationships. 
Besides, removing the emotion-guided domain mapping led to a decline in performance.
This implies that identifying the source domain and the target domain as well as tracing the emotional connections between the two domains can enhance the model's ability.
Furthermore, the performance declined when the verification mechanism is removed, which indicates that re-evaluate the results contributes to the improvement of performance.

\begin{table}[t!]
\centering
\resizebox{0.46\textwidth}{!}{
\renewcommand{\arraystretch}{0.9} 
\begin{tabular}{lcccc}
\toprule
& \multicolumn{2}{c}{\textbf{HYPO}} & \multicolumn{2}{c}{\textbf{HYPO-L}} \\
\cmidrule(lr){2-3} \cmidrule(lr){4-5}
& \textbf{Hyperbole} & \textbf{Metaphor} & \textbf{Hyperbole} & \textbf{Metaphor} \\
\midrule
Ours & 90.8 & 84.5 & 79.3 & 80.3 \\
\midrule
\quad w/o emotion & 86.2 \scriptsize{(-4.6)} & 79.4 \scriptsize{(-5.1)} & 74.7 \scriptsize{(-4.6)} & 74.6 \scriptsize{(-5.7)} \\
\quad w/o interaction & 87.4 \scriptsize{(-3.4)} & 80.7 \scriptsize{(-3.8)} & 75.8 \scriptsize{(-3.5)} & 75.3 \scriptsize{(-5.0)} \\
\quad w/o domain & 88.2 \scriptsize{(-2.6)} & 81.2 \scriptsize{(-3.3)} & 76.6 \scriptsize{(-2.7)} & 76.0 \scriptsize{(-4.3)} \\
\quad w/o verification & 89.3 \scriptsize{(-1.5)} & 83.1 \scriptsize{(-1.4)} & 78.1 \scriptsize{(-1.2)} & 78.4 \scriptsize{(-1.9)} \\
\midrule 
& \multicolumn{2}{c}{\textbf{LCC}} & \multicolumn{2}{c}{\textbf{TroFi}} \\
\cmidrule(lr){2-3} \cmidrule(lr){4-5}
& \textbf{Hyperbole} & \textbf{Metaphor} & \textbf{Hyperbole} & \textbf{Metaphor} \\
\midrule
Ours & 84.9 & 91.3 & 84.2 & 76.6 \\
\midrule
\quad w/o emotion & 79.6 \scriptsize{(-5.3)} & 85.9 \scriptsize{(-5.4)} & 79.8 \scriptsize{(-4.4)} & 72.2 \scriptsize{(-4.4)} \\
\quad w/o interaction & 80.9 \scriptsize{(-4.0)} & 87.2 \scriptsize{(-4.1)} & 80.6 \scriptsize{(-3.6)} & 73.6 \scriptsize{(-3.0)} \\
\quad w/o domain & 81.4 \scriptsize{(-3.5)} & 88.1 \scriptsize{(-3.2)} & 81.4 \scriptsize{(-2.8)} & 73.9 \scriptsize{(-2.7)} \\
\quad w/o verification & 83.4 \scriptsize{(-1.5)} & 89.9 \scriptsize{(-1.4)} & 82.9 \scriptsize{(-1.3)} & 75.2 \scriptsize{(-1.4)} \\
\bottomrule
\end{tabular}
}
\caption{Ablation results. The numbers in the brackets are
the decreased values compared with our full model.}
\label{tab:ablation}
\vspace{-4mm} 
\end{table}

\subsection{Discussion}
To further investigate the effectiveness of our method, we conduct in-depth analyses to answer the following questions, aiming to deeply mine the intuition and analyze implicit phenomena.

\begin{figure}[!t]
    \centering
    \includegraphics[scale=0.5]{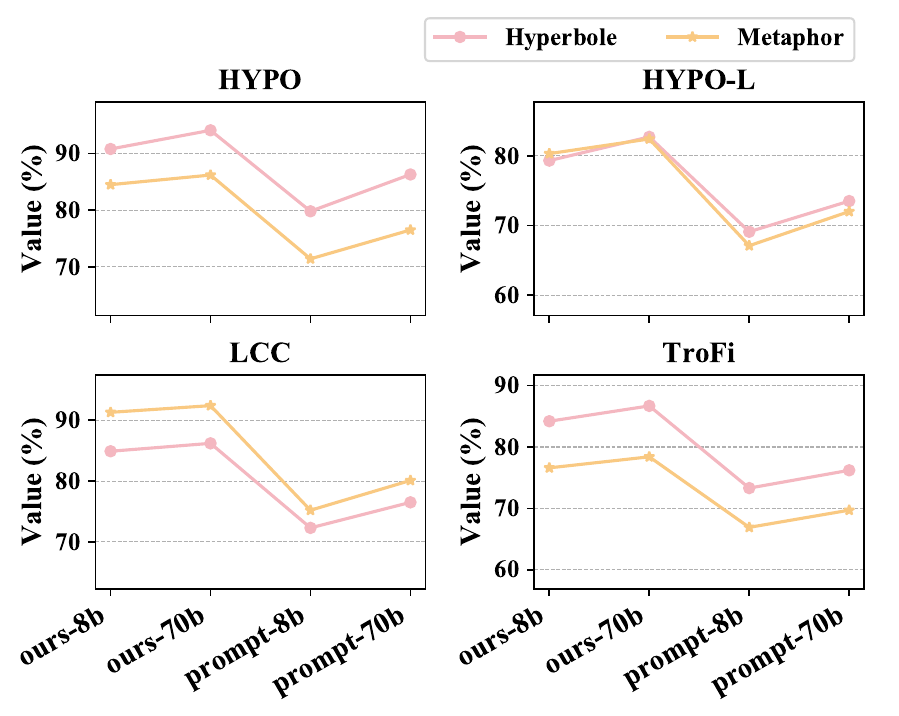}
    \vspace{-2mm}
    \caption{Comparison results of different LLM scales.}
    \label{fig:llm-scale}
    
\end{figure}

\noindent\textbf{1) What are the impacts of LLM scales?}
With the attempt to investigate the impact of different LLM scales, we evaluate the hyperbole and metaphor detection performance of Llama models with different sizes on four datasets.
As illustrated in the Figure \ref{fig:llm-scale}, it can be observed that for both our method and the prompt-based method, the performance of hyperbole and metaphor detection improves as the model scale increases. Moreover, we discover that compared with the prompt-based method, our method exhibits a more significant enhancement in performance when the model scale is enlarged. 
This indicates that our method can deeply mine the semantic and emotional information within the text, thereby demonstrating stronger adaptability in hyperbole and metaphor detection tasks.

\begin{figure}[!t]
    \centering
    \includegraphics[scale=0.7]{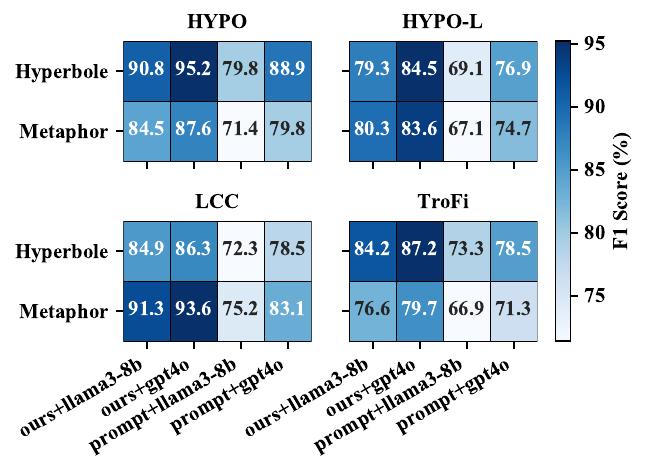}
    \vspace{-2mm}
    \caption{Comparison results of different LLMs.
    }
    \label{fig:llm-diff}
    \vspace{-0.6cm}
\end{figure}

\noindent\textbf{2) What are the influences of different LLMs?}
To explore the impact of different LLMs on hyperbole and metaphor detection, we select two representative models: Llama3-8b and GPT-4o for comparative experiments.
The experimental results are shown in Figure \ref{fig:llm-diff}. 
We observe that on all four datasets, GPT-4o consistently outperforms Llama3-8b. 
This consistent performance gap indicates that more powerful LLMs indeed have better rhetorical comprehension abilities. 
Additionally, we find that the performance of our method far surpasses that of prompt-based method, regardless of whether it is applied to Llama3-8b or GPT-4o. 
This shows that by deeply mining the semantic and emotional cues in the text, we compensate for the potential deficiencies of LLMs in handling rhetorical detection tasks, proving the effectiveness of our method.

\begin{figure}[!t]
    \centering
    \includegraphics[scale=0.3]{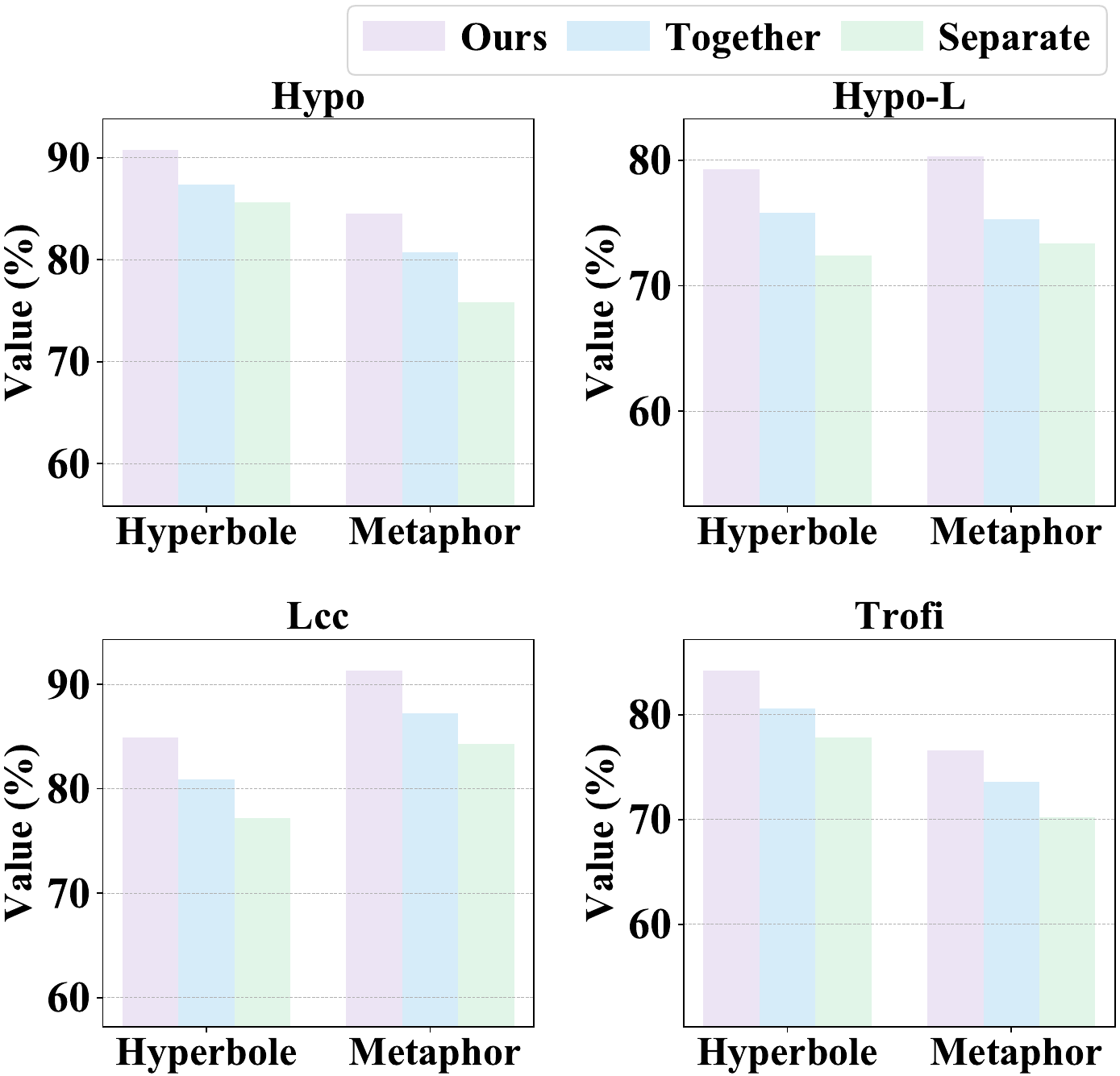}
    \vspace{-2mm}
    \caption{Influence of the bidirectional dynamic interaction mechanism in hyperbole and metaphor detection.
    }
    \label{fig:bi}
    \vspace{-0.6cm}
\end{figure}

\noindent\textbf{3) What are the advantages of the bidirectional dynamic interaction mechanism?}
We are curious about the effectiveness of the bidirectional dynamic interaction mechanism.
In Figure \ref{fig:bi}, we design three groups of experiments for comparison, namely: 1) \textit{detecting hyperboles or metaphors separately (Separate).}
2) \textit{detecting hyperboles and metaphors simultaneously (Together).} 
3) \textit{detecting hyperboles and metaphors using the bidirectional dynamic mechanism (Ours).} 
We observe that our method outperforms both separate and simultaneous detection on the four datasets. 
This indicates that the mechanism can better capture the semantic features.
It averts information one-sidedness in separate detection and overcomes interaction deficiency in simultaneous detection, yielding more accurate and efficient results. 
Moreover, we find that the performance of separate detection is the worst, which proves that hyperboles and metaphors can help each other and highlights the importance of integrating the effective knowledge between them.

\subsection{Case Study}
\begin{figure}[!t]
    \centering
    \includegraphics[scale=0.35]{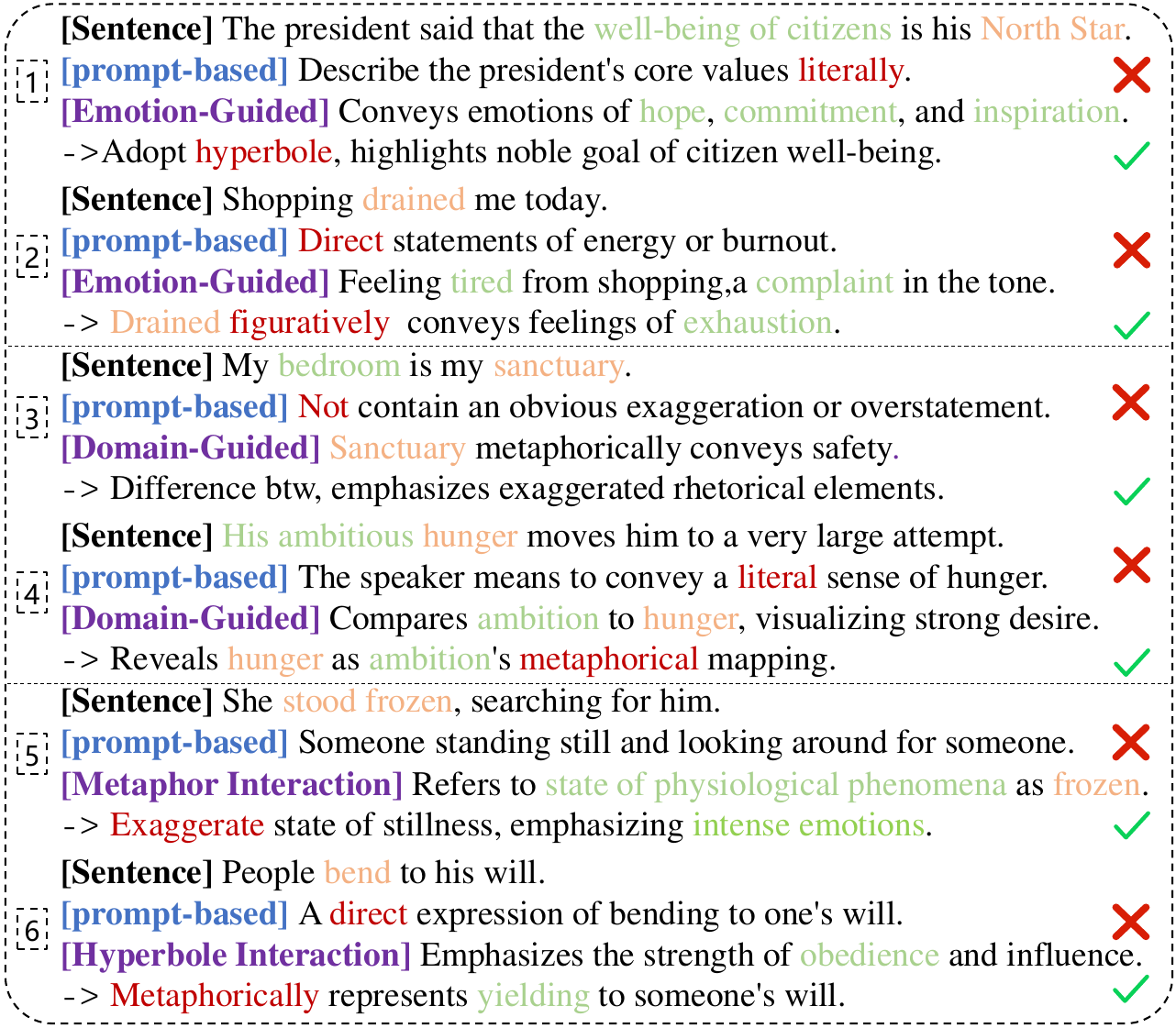}
    \vspace{-2mm}
    \caption{Case study to compare with prompt-based and our framework.
    }
    \label{fig:case}
    \vspace{-0.6cm}
\end{figure}

We conduct a case study to gain a deeper understanding of the importance and effectiveness of our framework.
As shown in Figure \ref{fig:case}, our framework successfully detects hyperboles and metaphors, while the prompt-based method fails. 
\circleone{1} In terms of emotion analysis, in Eg.1, our method correctly detects it as a hyperbole based on the emotions of hope and commitment it conveyed, yet the prompt-based method misses this. 
Similarly, in Eg.2, guided by the shopping tired emotion, our method successfully detects the sentence as hyperbole. 
In contrast, the prompt-based method misinterprets it as a direct statement.
\circletwo{2} Regarding the emotion-based domain mapping, in Eg.3, our method discerns the source domain ``sanctuary'' and the metaphorical meaning of safety and comfort, while the prompt-based method fails to recognize it as an exaggeration. 
In Eg.4, our method pinpoints ``hunger'' as the source domain and ``ambitious'' as the target domain, likening ambition to hunger to express a strong desire. 
Whereas the prompt-based method erroneously views the sentence as simply expressing hunger.
\circlethree{3} 
Concerning the bidirectional dynamic interaction between hyperbole and metaphor, in Eg.5, our method analyzes the metaphor that a person's body is frozen like a statue and thus infers that the sentence exaggerates the static state. 
The prompt-based method, unfortunately, fails to identify the hyperbole.
In Eg.6, Our method reasons out the metaphorical nature of ``bend'' by identifying the hyperbole of his influence in the sentence, which the prompt-based method cannot capture.
Overall, this analysis emphasizes the significant meaning and effectiveness of the emotion analysis, the emotion-based domain mapping, and the bidirectional dynamic interaction between hyperbole and metaphor in precisely detecting hyperboles and metaphors.

%% file: Section/5-conclusion.tex
\section{Conclusion}

In this paper, we proposes an emotion-guided hyperbole and metaphor detection framework based on bidirectional dynamic interaction(EmoBi). 
By means of emotion analysis, emotion-based domain mapping, and bidirectional dynamic interaction mechanism, it fully utilizes the interaction between emotion information and tasks to enhance the detection performance. 
Through in-depth analysis, it is discovered that EmoBi can compensate for the deficiencies of LLM in handling specific rhetorical detection tasks by mining emotion cues. 
The experimental results on four widely-used datasets demonstrate the effectiveness of our proposed innovative method, achieving SoTA performance.

\section{Limitations}
Despite the remarkable achievements of the proposed EmoBi in this paper, there are still some limitations that present opportunities for further improvement.
First, due to its multi-step reasoning approach, EmoBi suffers from the issue of error propagation. When an error occurs in the previous steps, it may affect the judgments in the subsequent steps. 
Second, while emotion knowledge plays a crucial role in our model, the current emotion analysis module may not always accurately capture all the subtle emotions. 
In future work, we will consider how to further ensure the quality of emotion knowledge to better assist in hyperbole and metaphor detection. 

\section*{Acknowledgments}

This work is supported by the National Key Research and Development Program of China (No. 2022YFB3103602), the National Natural Science Foundation of China (No. 62176187). 